\documentclass[letter, 10 pt, conference]{ieeeconf}
\IEEEoverridecommandlockouts                            
\usepackage{amsmath,amsfonts}
\usepackage{algorithmic}
\usepackage{algorithm}
\usepackage{array}
\usepackage[caption=false,font=normalsize,labelfont=sf,textfont=sf]{subfig}
\usepackage{textcomp}
\usepackage{stfloats}
\usepackage{url}
\usepackage{verbatim}
\usepackage{graphicx}
\usepackage{cite}
\usepackage{makecell}
\usepackage[table]{xcolor}
\usepackage{multirow}
\usepackage{siunitx}
\usepackage{etoolbox}
\usepackage{multirow}
\usepackage{array}
\usepackage[T1]{fontenc}
\usepackage{lipsum}

\begin{document}

\title{\LARGE \bf 3D Printable Soft Liquid Metal Sensors for Delicate Manipulation Tasks}

\author{Lois Liow,~\IEEEmembership{Member,~IEEE},
   Jonty Milford,
   Emre Uygun,
   André Farinha, \\
   Vinoth Viswanathan,
   Josh Pinskier,
   and David Howard,~\IEEEmembership{Senior Member,~IEEE}%
\thanks{L. Liow, E. Uygun, A. Farinha, V. Viswanathan, J. Pinskier, and D. Howard are with CSIRO Robotics, Pullenvale, QLD 4069, Australia. All correspondence should be addressed to {\tt\small lois.liow@csiro.au}}%
\thanks{J. Milford is with Flinders University, Bedford Park SA 5042, Australia.}}

        % <-this % stops a space

\maketitle
\thispagestyle{empty}
\pagestyle{empty}

%%%%%%%%%%%%%%%%%%%%%%%%%%%%%%%%%%%%%%%%%%%%%%%%%%%%%%%%%%%%%%%%%%%%%%%%%%%%%%%%

%%%  Only 2000 characters (including whitespaces) allowed! %%%
\begin{abstract} 

Robotics and automation are key enablers to increase throughput in ongoing conservation efforts across various threatened ecosystems.  Cataloguing, digitisation, husbandry, and similar activities require the ability to interact with delicate, fragile samples without damaging them.  Additionally, learning-based solutions to these tasks require the ability to safely acquire data to train manipulation policies through, e.g., reinforcement learning.  To address these twin needs, we introduce a novel method to print free-form, highly sensorised soft ‘physical twins’. We present an automated design workflow to create complex and customisable 3D soft sensing structures on demand from 3D scans or models. Compared to the state of the art, our soft liquid metal sensors faithfully recreate complex natural geometries and display excellent sensing properties suitable for validating performance in delicate manipulation tasks.  We demonstrate the application of our physical twins as 'sensing corals': high-fidelity, 3D printed replicas of scanned corals that eliminate the need for live coral experimentation, whilst increasing data quality, offering an ethical and scalable pathway for advancing autonomous coral handling and soft manipulation broadly. Through extensive bench-top manipulation and underwater grasping experiments, we show that our sensing coral is able to detect grasps under 0.5\,N, effectively capturing the delicate interactions and light contact forces required for coral handling. Finally, we showcase the value of our physical twins across two demonstrations: (i) automated coral labelling for lab identification and (ii) robotic coral aquaculture. Sensing physical twins such as ours can provide richer grasping feedback than conventional sensors providing experimental validation of prior to deployment in handling fragile and delicate items.

\end{abstract}

%%%%%%%%%%%%%%%%%%%%%%%%%%%%%%%%%%%%%%%%%%%%%%%%%%%%%%%%%%%%%%%%%%%%%%%%%%%%%%%%
\section{INTRODUCTION}

\begin{figure*}[t]
\centerline{\includegraphics[width=1\linewidth]{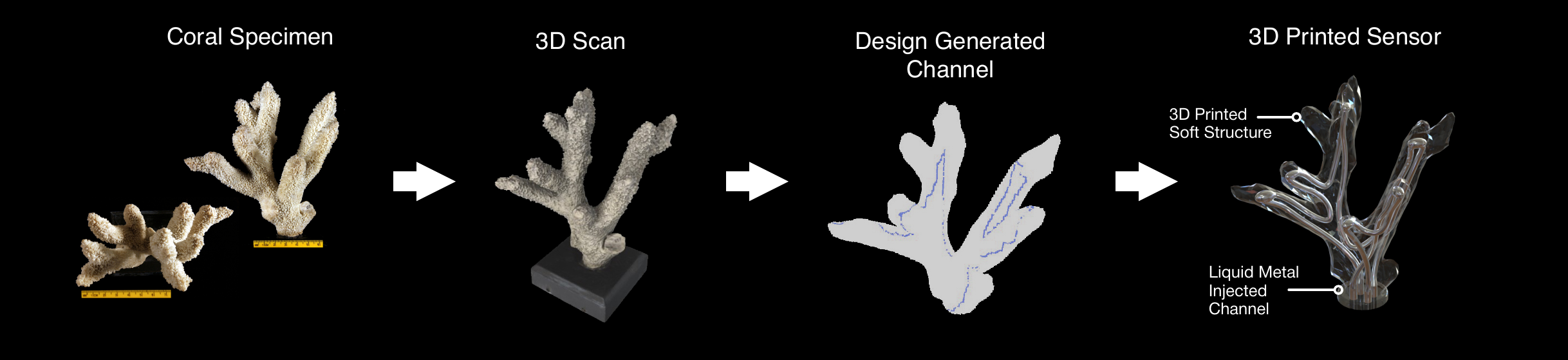}}
\caption{Left to right: \textit{Madrepora robusta (Dana, 1846)} coral \cite{dana1846zoophytes}, 3D scan of the coral \cite{dana1846zoophytes}, mesh converted to a voxelised model with an automated design generated channel, and a printable model of the coral for LM sensing. Coral images and 3D scans are available from the Smithsonian Institution \cite{dana1846zoophytes}.}
\label{fig:generation}
\end{figure*}

Robotic manipulation is a rapidly advancing field, whose growth has recently been accelerated by the creation of new, capable dexterous manipulation hardware, and generative AI models  \cite{zhang2024affordance,chi2023diffusion}. Combined, these tools provide the basis for generalised, safe robotic handling and manipulation which is starting to be evidenced in household and laboratory settings. However, a range of high value applications in healthcare, agriculture, and ecology are currently overlooked because of the difficulty and risk involved in testing systems and obtaining data of fragile (potentially living) samples.  These specimens provide a range of unique challenges---they are geometrically diverse, may be soft or compressible, and may be delicate and fragile, tolerating only very low forces.  To realise impact in these areas, we require a way of safely prototyping systems that interact with delicate objects.

As an example, consider the handling of coral for applications in digitisation and husbandry.  Coral are a diverse class of creatures with widely diverging geometric and mechanical properties\cite{zawada2019quantifying}. They can be hard or soft, branching or encrusting and range in size from millimetres to metres\cite{chandler2024predicting}. Despite this diversity, all coral share the trait of being extremely fragile and highly susceptible to stress and damage from improper handling. For robots to effectively perform delicate tasks such as coral handling, it is essential for researchers to first develop an understanding of the forces experienced by the target samples. The majority of current research focuses on sensors attached to the manipulator. However, for delicate samples, a more object-centric approach is suggested---particularly one that guarantees measurement around these exerted forces, which is a critical step towards real-world deployment.

In soft manipulation, prior work includes vision-based deformation estimation using depth cameras to measure their global deformation \cite{greenland2025sograb}. However, these methods are sensitive to camera occlusion and not suitable for all applications---specifically in-situ applications with limited space precluding the use of a camera.  Deformation can also be measured locally through contact, via in-hand embedded tactile sensors or cameras \cite{luu2025vision,khalid2025camera, bui2024rose,yuan2017gelsight}. However, this method only works for simple geometries, such as cylinders, and is not suitable for structures with irregular, organic morphologies.

Closest to our work is the use of sensorised objects \cite{10538419} whose morphologies are designed to mimic objects in the real world, and which provide a more complete picture of the forces experienced by the object being manipulated. For example, a sensorised soft raspberry replicates the physical interactions and mechanical response of a real raspberry, and is used to prototype harvesting systems \cite{junge2023lab2field}. Robotic Phantoms for remote medical training provide another prominent example, as they are able to simulate a variety of physiological conditions through variable stiffness mechanisms \cite{he2020abdominal}. However, existing physical twins are limited to a simple morphologies, and no method currently exists to generate free-form  sensorised soft objects.

% Deformable object manipulation requires the selection of appropriate sensors to obtain meaningful object representation, and most often, a fusion of different sensing modalities are required \cite{zhu2022challenges}. To reconstruct an object's deformed state, researchers have used a combination of vision, tactile sensing on the gripper, and even sensorising the deformable object itself-- and creating "physical twins". 

In this paper, we propose an automated framework to create arbitrarily shaped, 3D printable, soft sensorised objects with embedded Liquid Metal (LM) channels, designed to capture delicate interactions in soft manipulation tasks, such as the handling of fragile corals. Automated design has long been used as a tool to rapidly create bespoke soft robots to fulfil demanding tasks and incorporate manufacturing requirements\cite{pinskier2022bioinspiration,pinskier2024diversity,tapia2020makesense}. Here it streamlines the design process for a range of geometrically complex shapes and minimises the need for human intervention, improving customisability and reducing the labour needed to fabricate these sensors. We demonstrate the utility of our method in marine conservation through the design of a sensing coral, developed as a soft physical twin that provides deformation feedback and training data to robotic manipulation systems. Using scans of real corals and their meshes, we introduce a design pipeline to: (1) automatically generate a hollow channel throughout the given 3D structure, (2) monolithically 3D print the soft object, and (3) fill the hollow channels with low-melting point LM for resistive sensing, as illustrated in Fig. \ref{fig:generation}. Low melting point LMs, such as Galinstan or eutectic gallium-indium (EGaIn) are recently emerging field of functional materials that are used to create soft robots, flexible robotic skins, and wearables \cite{ma2023shaping}. Whilst substantial existing research exists into their fabrication onto 2D planes \cite{ma2023shaping}, their development into 3D structures has received little attention. Where they have been developed, it required cumbersome post-processing though dissolvable sacrificial wax \cite{bharambe2017vacuum}. To address this problem, we present an efficient 3D printing solution which uses Stereolithography (SLA) printing to manufacture free-form 3D structures without need for internal support or post-processing. The proposed method is a promising pathway to create more customisable soft sensing systems on demand for applications such as high-throughput experimentation. 

The rest of the paper is organised as follows: Section II details the design workflow used to generate the LM channel paths for a given 3D mesh structure (available from the Smithsonian Institution  \cite{dana1846zoophytes}), the 3D printing fabrication technique, the LM injection process, and the resistive measurement setup used for data collection. Section III discusses the sensor characterisation  and grasping experiments to determine the sensitivity of the coral across each of its sensing branches, and its ability to assist with regulating gripper forces in a pick-and-place task based on human guidance. Section IV serves as a demonstration to show possible use cases for the sensing coral. Sensor feedback from the coral is used to detect critical events, such as overtightening of the wire when attaching label to a coral. To validate the performance of the sensing coral in submerged environments, experiments with a gantry type coral farming robot were also conducted. Finally, Section V concludes the article with suggestions for future work.

Coral reefs are among the most fragile and yet the most vital ecosystems on the planet, currently under severe threat from climate change and human activity. There is a need for non-invasive robotic technologies to assist with ongoing reef restoration activities. Hence, our soft sensors were carefully designed to be waterproof and non-toxic for marine lifeforms, as the 3D printing material was carefully selected to be safe for underwater testing. Corals also present a unique challenge for robotics, as their complex irregular geometries and branched structures make them difficult for robots to manipulate them safely. Traditional gripper-based sensors may struggle to capture the nuanced contact interactions that occur along these structures, and risk damaging the living coral. By developing these soft physical twin corals, we create a safe testing ground to train robotic manipulation systems, that can eventually be used to handle live corals. 

\begin{figure*}[t]
\centerline{\includegraphics[width=0.9\linewidth]{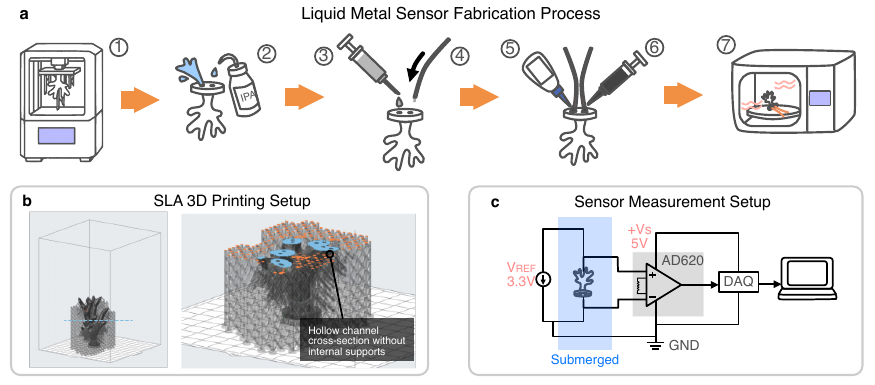}}
\caption{(a) Fabrication process of soft sensing coral. (b) 3D printing setup with no internal supports and channel cross-section checked along z-axis to ensure no excessive overhangs. (c) Measurement setup designed for low resistance measurement and submersion. }
\label{fig:fab}
\end{figure*}

\section{Sensor Design and Fabrication}
\label{section:method}
\subsection{Automated Sensing Channel Generation}
To fabricate the sensors, a valid channel pathway was identified for each scanned coral sample. The channel is required to be continuous with a non-intersecting path, to span the entire coral sample including its branches, and must be suitable for 3D printing with no large overhangs of unsupported material. This path planning problem was approached with the ideal channel being defined as the shortest, non-intersecting path that passes through a set of user-defined keypoints, which was then solved using Dijkstra's shortest path algorithm. The process consisted of the following steps: (1) a 3D scanned coral mesh was first  voxelised to generate a binary 3D array represented by voxels on the coral's surface; (2) a set of keypoints which the path should pass through was then defined by the user, with the first and last point in the base of the coral; (3) using the 3D array as a graph, the shortest path was found iteratively using Dijkstra'a algorithm \cite{dijkstra3d}, to find paths between adjacent keypoints one segment at a time. Dijkstra's method was used to find the shortest path along a weighted set of graph nodes (or voxels in the 3D implementation). To prevent self-intersection in subsequent segments, the identified path segments were assigned with an arbitrarily high cost (\num{1e6} versus \num{1} for unoccupied paths).
After all segments have been completed, the result was a continuous path which starts and ends at the coral trunk. 
A strength of this method is its flexibility and user configurability, which allow it to seamlessly transfer to different 3D printing methods (with their own requirements) and coral geometries.
Path constraints (e.g limiting travel directions) and voxel weightings (e.g to bias towards the interior of the coral rather than surface) can be added into the search method by changing the optimisation function or voxel weights, to configure the solution to user requirements and preferences.

\subsection{Material Selection and 3D Printing Process}

The soft sensor was 3D printed using Formlabs Silicone 40A material with a Formlabs Form 3+ SLA printer, with the full fabrication process illustrated in Fig.\ref{fig:fab} (a). Formlabs Silicone 40A was selected due to the low resin viscosity, low stiffness (Shore 40A), and high durability of the parts after printing. Our proposed fabrication technique, is not limited to SLA 3D printing and this particular resin material. Recommended materials and best practice guidelines are provided to researchers at the end of this section. 

For underwater applications, suitable 3D printing methods are substantially limited as the printed parts need to be watertight and non-absorbent, which effectively rules out the fused deposition modelling (FDM) method that results in inherently porous structures. In contrast, SLA and Digital Light Processing (DLP) methods have been traditionally used to create watertight enclosures. Silicone 40A material was chosen because it is not absorbent to water, with \textless0.1\% weight gain after being immersed in both distilled and salt water for 24 hours, as reported by the supplier's datasheet.

In the print slicer, the coral model was orientated upright, with its base placed flat against the bed, as in Fig. \ref{fig:fab}(b). This orientation was chosen to avoid the channel being printed with large overhangs, which may lead to print failure. To check for overhangs, the model was sectioned across different heights along the z-axis to ensure that the cross-section of the channel was roughly circular in shape. This ensures the subsequent layers are well-supported by the previous layers. One of the benefits of this 3D printing method is that internal supports for the channel is not necessary, therefore reducing the time and effort required for post-processing. Once the print is completed, the external supports were removed and the print was washed with a mixture of butyl acetate and isopropyl alcohol (IPA), and the channel was subsequently flushed with IPA until all resin was removed. This is an important step in the process to ensure that the channel does not become blocked if there is any trapped residual resin. Once clean, the print was submerged in a water bath and cured under UV light with heat. The cured print was then injected with LM with a syringe into the channel inlet hole until LM starts to flow out from the outlet. The LM used for the experiments was pure Eutectic Ga-In-Sn (Zinn Giesserei Goehler), otherwise known as Galinstan, with a composition: 68.5\% Ga, 21.5\% In and 10\% Sn. The melting point of the LM is \SI{11}{\degreeCelsius} and density is \SI{6.44}{g/cm³}. Compared to other conductive composites which contain conductive particles within a non-conductive fluid (e.g carbon or silver grease), homogeneous liquid metals exhibit greater linearity, lower drift and hysteresis.
% In initial experiments using carbon- and silver-based greases and composites, resistance measurements exhibited noticeable drift due to hysteresis after the structures were compressed and then released. This phenomenon occurs because the conductive filler particles in a non-conductive fluid medium do not perfectly reform conductive pathways when the material is deformed and subsequently returns to its original shape. For this reason, LM was selected as the most appropriate conductive fluid for our application, as it exhibits high linearity and low hysteresis.
Once the channel was filled with LM, wires were inserted into the inlet and outlet openings and were adhered in place temporarily with cyanoacrylate glue. The same SLA resin, Silicone 40A was then injected around base of the coral with a syringe with a large blunt tip nozzle to complete the sealing. Finally, the coral was fully cured together with the wires and is ready to be used. Each coral sensor was produced with \SI{11}{\gram} of Galinstan (USD \$18) and \SI{78}{\milli\liter} of Silicone 40A resin (USD \$36).

Through iterative testing, the following considerations and guidelines are proposed to assist researchers in fabricating these sensors. It is recommended to select SLA or DLP resin with low uncured viscosity. Extremely viscous resins were found to be unsuitable for printing, as their adhesive behaviour causes the resin to accumulate and cure within the channel during the printing process. As a result, the channel may become obstructed even before the print is completed. Another consideration is shrinkage of the material after post-curing, which will reduce the size of the printed channel compared to what was originally designed. Although the channel may be clear during the IPA cleaning process, it may become blocked after UV curing. Hence, to summarise, it is highly recommended for researchers to select resins with low viscosities and low shrinkage after curing (or increase diameter of the channel to account for shrinkage) to ensure successful prints.

\begin{figure*}[t]
\centerline{\includegraphics[width=0.95\linewidth]{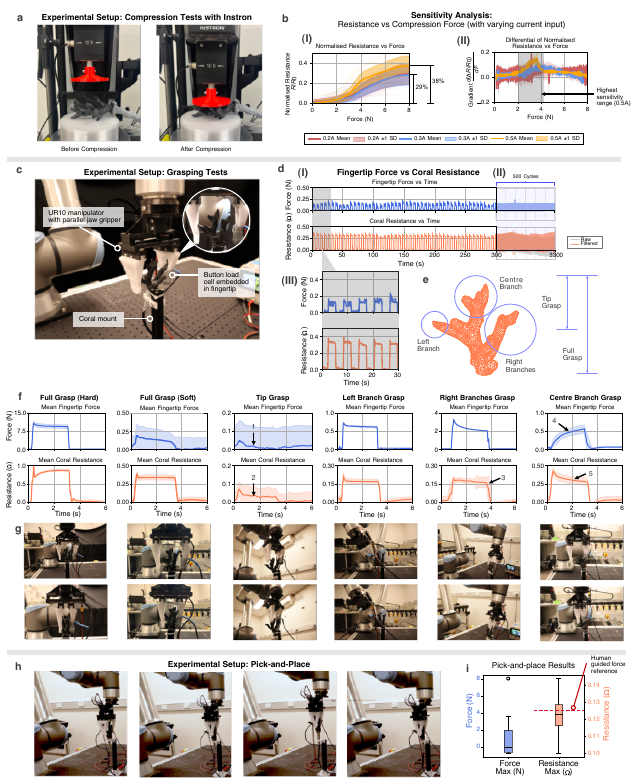}}
\caption{(a) Instron compression experiments. (b)(I) Normalised coral resistance vs force plot showing greater sensitivity with increase in current. (II) Gradient of normalised resistance vs force plot. (c) Grasping experiments with a gripper and embedded load cell on a fingertip, mounted on a UR10 robotic manipulator for positioning. (d)(I) Fingertip force and coral resistance over time for 50 cycles. (II) Baseline drift of resistance measurements over 500 cycles. (III) First \SI{30}{\second} of fingertip force and coral resistance vs time. (e) Identification of the coral branches. (f-g) Manipulator repositioned to grasp various branches, recording fingertip forces and corresponding resistance over time. Results reported were the mean and SD across 50 cycles. Callout 1-2: Fingertip forces quickly deteriorated, while the coral's resistance provided consistent detection of the soft contact. 3: The load cell only measured contact with the nearest branch, whereas the coral's resistance allowed the detection of the neighbouring branch (shown by the shaded area). 4-5: Contact with the three-pronged centre branch was detected by the load cell only after sufficient compression, whereas the sensing coral provided immediate feedback upon contact. (h) Pick-and-place experiments. (i) Box-and-whiskers plot of maximum fingertip forces and resistance values across 15 experiments.}
\label{fig:experiments}
\end{figure*}

\subsection{Design of Resistance Measurement Setup}

% The resistance of the sensing coral was found to be \SI{0.5}{\ohm} without deformation. 
As LM's exhibit low resistance, a precise measurement method was required. The 4-wire Kelvin method was selected over the Wheatstone bridge, due to simplicity of the setup. One of the disadvantages of the Wheatstone bridge is that the bridge becomes inherently non-linear further away from the balance point. In our application, we have observed exceptionally large changes in resistance after deformation and therefore opted for the 4-wire method for greater accuracy and its linear behaviour, which is dependent on the current input. A constant current of \SI{0.2}{A} and voltage limit of \SI{3.3}{V} was applied across the coral using a bench-top power supply. The voltage drop across the coral was amplified using an AD620 instrumentation amplifier (gain 1.5) before it was read using a Data Acquisition System (DAQ, National Instruments USB-6211). The high impedance of the op-amp inputs eliminates the effect of the sense leads' resistances, allowing long cables to be used. For our underwater experiments, described in Section. \ref{section:demo}, a 2-pair twisted shielded cable of \SI{1.2}{\meter} was required to route the cable out of the water tank, as shown in Fig. \ref{fig:fab} (c). The base of the coral, where the PVC sleeve terminates is sealed with room-temperature vulcanising (RTV) silicone to ensure that water does not enter the cable.  

% Although touch capacitive sensing was considered due to the high conductivity of the LM, it is not suitable to be used in underwater environments, especially salt water, due to its conductive nature and hence shielding would be required. Furthermore, capacitance sensing requires two conductive layers, which would require more channels and thus, increases complexity of the sensor fabrication. 

\section{Sensor Performance Evaluation}
\label{section:experiments}
\subsection{Sensitivity Characterisation Experiments}
\textbf{Compression Experimental Method: }The sensing coral was compressed using a Universal Testing System (Instron 34SC-5 Single Column Table) with a \SI{10}{\newton} limit load cell. The coral was orientated horizontally between the compression plates, as in Fig. \ref{fig:experiments}(a), and increasing force was applied until \SI{8}{\newton} was reached. The sensitivity of the coral was evaluated by measuring the increase in resistance over the applied range of compressive force. The sensitivity of the coral was found to increase with the constant current and in this experiment, the sensor's response was evaluated for \SI{0.2}{A}, \SI{0.3}{A}, and \SI{0.5}{A} of current. For the three sets of experiments, the coral was compressed for 20 cycles each and the mean change in resistance and corresponding standard deviation (SD) were calculated. 

\textbf{Compression Experimental Results: }In Fig. \ref{fig:experiments} (b)(I), when the current through the coral was increased, a greater change in resistance was observed for the same applied force. The results reported were the coral's normalised resistance, which is defined as the change in the coral's resistance ($\Delta R$) relative to its initial baseline resistance without deformation ($\Delta R_0$). Hence, the baseline resistance of \SI{0.5}{\ohm} was subtracted to zero the measurements before data is collected. For \SI{0.2}{A} and \SI{0.3}{A}, the coral's resistance was found to increase up to approximately 29\% of the full range. In contrast, when the current was increased to \SI{0.5}{A}, the resistance increases by about 38\% of the full range. This range percentage  is an important consideration to ensure that the sensor designed is sensitive enough for the range of forces unique to the application. With \SI{0.5}{A}, a steeper gradient between \SIrange{2}{4}{\newton} was also observed and this is highlighted in Fig. \ref{fig:experiments} (b)(II) with the differential of the normalised resistance to the applied force. Hence, the coral was the most sensitive between \SIrange{2}{4}{N}, offering improved resolution for measuring smaller relative resistance changes within this range. For the remaining experiments, a constant current of \SI{0.2}{A} was maintained for consistency and to keep the system low-powered for underwater experiments.

\subsection{Sensor Performance Experiments}
\textbf{Grasping Experimental Method:} The sensing coral was mounted in an elevated position, as in Fig. 
\ref{fig:experiments}(c). A custom parallel jaw gripper was developed and attached to the end-effector of a Universal Robots UR10 robotic manipulator. The gripper consists of two rigid fingers, actuated with a Dynamixel motor (ROBOTIS XM430-W210) with a linear slider mechanism. A button compression load cell (TE Connectivity FX293X-100A-0010-L) with a \SI{45}{\newton} limit was used to measure the fingertip contact forces. The load cell was casted with flexible potting epoxy to ensure it was waterproof for the subsequent submerged experiments. The coral was grasped in repetition, and corresponding fingertip forces provided by the load cell and the coral's change in resistance were measured with the DAQ at \SI{1000}{Hz} sampling rate. For each grasping cycle, the gripper was opened for \SI{3}{\second} and then closed for \SI{3}{\second}. The plot in Fig. \ref{fig:experiments}(d)(I) shows the fingertip force and change in coral resistance over time across 50 cycles. To characterise the stability of the sensor in the long-term, 500 cycles was subsequently performed to measure the baseline drift, as in Fig. \ref{fig:experiments} (d)(II). The performance of each of the individual and grouped branches were also evaluated. The gripper was repositioned using the manipulator to access the different branches, as in Fig. \ref{fig:experiments} (e-g). The reported results are the computed mean and SD across 50 cycles for each set of experiments. In the experiments, different grasp types were classified based on the closing distance of the gripper. For 'hard grasps', the gripper was actuated until the fingers were \SI{1}{\centi\meter} apart, while for 'soft grasps' the fingers were \SI{3}{\centi\meter} apart, and for grasping individual or grouped branches, the gripper was driven until it reached its maximum closure.

To evaluate the reliability of the sensing coral for robotic manipulation applications, pick-and-place experiments were also conducted, as in Fig. \ref{fig:experiments} (h). The coral's resistance was used as feedback to autonomously regulate the grasping force of the gripper. A human-guided force reference was first established to define the target forces needed to pick up the coral. This was done by measuring the resistance reported by the coral as a person picks it up from its mount, which corresponded to a peak resistance of \SI{0.125}{\ohm}. The gripper was then programmed to close around the coral in steps until this target resistance was reached, in which the gripper stops completely. 15 pick-and-place cycles were performed and the maximum fingertip forces and coral resistances for each cycle were computed to produce a box-and-whiskers plot, as shown in Fig. \ref{fig:experiments}(i). It is important to note that human-coral contact interactions are far more complex than the simplified representation presented here and future work will investigate these intricate interactions to develop a deeper understanding of the bimanual coral handling task. 

\textbf{Grasping Experimental Results: }With reference to Fig. \ref{fig:experiments}(d)(I), the coral resistance measurements were found to be more consistent with less variability between cycles compared to the fingertip forces. Due to the irregular geometry and inherent flexibility of the coral, the branches tended to displace when grasped. Hence, the load cell was unable to provide an accurate reading in most cases as the contacting surface often moved away from its centre. In robotic manipulation studies, multi-taxel tactile sensors are often preferred to capture these dynamic contact interactions, but to the authors' knowledge, there are currently no tactile sensors IP rated for submersion that is purchasable off the shelf, which justifies our approach to sensorise the grasped object instead. Results have also shown good stability of the sensor after 500 grasping cycles, an hour of continuous experimentation. The raw resistance data and filtered data is presented in Fig. \ref{fig:experiments} (d)(II). Although a baseline drift over time was observed, this was easily removed by using a high-pass filter and by subtracting a linear least-squares fit of the baseline. The baseline drift in measurements was likely caused by the slight heating of the LM and wear of the soft printed material over time. Overall, the results showed excellent response times, repeatability, and low hysteresis. 

\begin{figure*}[t]
\centerline{\includegraphics[width=0.9\linewidth]{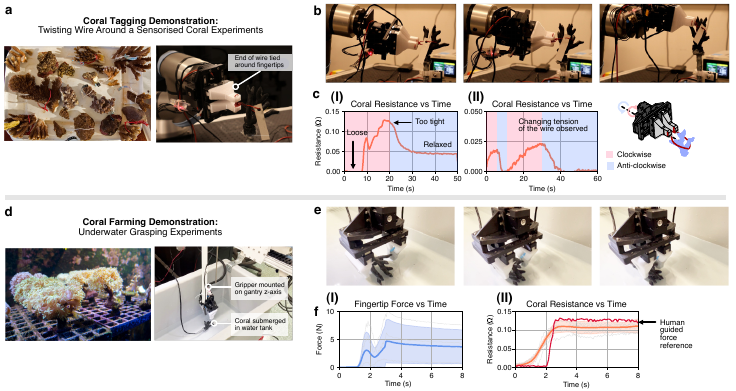}}
\caption{(a) Left: Corals tagged in the field (\textit{image source: Adobe Stock}). Right: Coral tagging demonstration setup. (b) The end-effector is rotated, twisting the wire to secure the tag against the coral. (c)(I) Example scenario: The wire is accidentally overtightened and then loosened to prevent excessive pressure on the coral. (II) Sensing coral detecting variations in wire tension with alternating rotation of gripper. (d) Left: Coral aquaculture (\textit{image source: Adobe Stock}). Right: Coral farming demonstration setup. (e) The submerged coral is grasped using resistance-based feedback control. (f)(I-II) Fingertip forces show large error regions, while the coral resistance increases towards the given target force reference.}
\label{fig:demonstration}
\end{figure*}

As discussed in the methods, the robotic manipulator was subsequently repositioned to grasp different branches on the coral, to evaluate their independent performance. A 'Full Grasp' was initially performed by using the gripper to envelop the entire coral up to the base. Hard grasps,  with mean peak fingertip forces of approximately \SI{11.1}{\newton}, were performed on the coral followed by significantly softer, more gentle grasps with forces of \SI{0.19}{N}. The resulting mean change in resistance is \SI{0.99}{\ohm} for the hard grasp and \SI{0.39}{\ohm} for the soft grasp. A distinguished peak was observed with the hard grasp, clearly capturing the instance when the fingers impact against the coral. This is also true for remaining experiments, where the coral resistance provided more distinguished plots with smaller error regions, compared to the load cell measurements. The difference in quality of the measurements were especially seen in experiments where softer grasps were performed, such as the 'Tip Grasp', whereby only the top section of the coral was gently pinched. Despite the low contact forces applied of under \SI{0.3}{\newton}, the coral was still able to detect grasps reliably. With reference to Fig. \ref{fig:experiments}(f), it can be observed that the mean fingertip force plot deteriorates over time (refer callout 1), while the resistance plot remained reasonably stable during the entire grasp duration (refer callout 2). Subsequent grasping experiments were also performed on different grouped and individual branches (refer Fig. \ref{fig:experiments}(e) for the classification of grasped regions). Grasping the singular left branch yielded mean fingertip forces of \SI{0.74}{\newton} and mean resistance of \SI{0.24}{\ohm}. As the gripper's relatively large size made it difficult to reach for individual branches without contacting neighbouring branches, the two rightmost branches were treated as single group, 'Right Branches Grasp', in this set of experiments. This grouping had created interesting results, whereby although the only a single force peak was measured by the load cell, but the coral's resistance feedback allowed detection of the neighbouring branch in some experimental cycles, as shown by the error region around the \SI{3}{\second} mark (refer callout 3). Another observation to note is that the 'Left Branch Grasp' and 'Right Branches Grasp' sets of experiments reported similar resistances values, just over \SI{0.2}{\ohm}, despite measuring vastly different fingetip contact forces (\SI{0.74}{\newton} compared to \SI{3.41}{\newton}). With the right branches, large fingertip forces were reported because two branches were grasped instead of one. Force from the gripper was used to deform the coral's soft material, rather than its LM channel. In contrast, although grasping a singular branch resulted in lower fingertip forces due to less material being compressed, the channel deformation is comparable to when two branches were grasped, and so the resistance values were similar for both cases. Finally, the 'Centre Branch Grasp' experiment yielded mean peak fingertip forces and coral resistances of \SI{0.62}{\newton} and \SI{0.46}{\ohm} respectively. The force plot differs from the rest of the experiments, in which it gradually rises until the \SI{3}{\second} mark, just before the gripper opens (refer callout 4). This inverted force plot is due to the irregular three-pronged geometry of the centre branch, as proper contact with the load cell only occurs after the coral is sufficiently compressed. On the other hand, the coral's resistance profile remains similar to previously reported plots, with a distinct peak immediately after contact (refer callout 5). Hence, the LM channel within the branches was still able to detect immediate contact even when the fingertip sensors fail to do so, thereby providing supplementary information about the interactions between the gripper and object. In summary, these key observations highlights the advantage of our sensing method, which provides distributed feedback across the entire structure, governed not by the body's morphology, but by the deformation of its internal LM channel, which can be seen as the body's nervous cord.

Finally, results of the pick-and-place experiments highlight the reliability of the sensing coal in a manipulation task. With feedback from the coral's resistance, the manipulator was able to perform 15 successful picks consecutively, with the distribution of maximum fingertip forces and resistance values shown in the box-and-whiskers plot in Fig. \ref{fig:experiments}(i). The median resistance was observed be below the human guided reference (indicated by the dotted red line), with the maximum (upper whisker) and minimum (lower whisker) values of \SI{0.143}{\ohm} and \SI{0.105}{\ohm} respectively. As the task was much more dynamic than static grasping experiments, the maximum forces measured by the load cell were much more disparate, with the box median around \SI{0}{\newton}, with a large upper quartile whisker and an outlier. This positive skew, indicates a large spread of contact forces measured, but in most experiments, these values were close to zero. This is because in most experiments contact with the coral did not occur at the fingertips, but rather, at the upper regions of the fingers due to the coral's spread-out branched structure. Beyond developing a printable sensor, our research highlights the challenges of understanding these complex interactions with irregular and organic structures that are especially common in nature, that cannot be captured by sensors on the gripper side alone.

\section{Demonstration}
\label{section:demo}
In this section, two demonstrations are provided to show the example use cases for the soft physical twin coral. The first focuses on one of the most time consuming tasks in the coral sampling process: coral tagging. Through interviews with marine scientists, we identified the need to automate this time-consuming, dull, and repetitive task. This task involves tying a wire or cable tie around a coral fragment with an identifying tag after they are collected in the field. These corals are subsequently transported back to the lab for further processing and preservation. One of the challenges of this task is that corals exhibit a diverse range of morphologies; some species, especially deep sea corals have very dense, intricate structures that are easily crushed. Further, there is a tendency for the tags to slip off if they are not attached properly. For the experimental setup, a wire was attached to the ends of two rigid fingertips on the same custom gripper that was previously used. The manipulator was first moved so that the wire hoops over the coral, before the fingers were subsequently closed to tighten the wire, as in Fig. \ref{fig:demonstration} (a-b). The manipulator's end-effector joint was then rotated continuously and the corresponding resistance of the coral was measured. In Fig. \ref{fig:demonstration}(c)(I), it can be observed for the first \SI{8}{\second} the wire remains loose and resistance reading is \SI{0}{\ohm}, but the resistance soon increases rapidly with the increasing tension of the wire. The tension of the wire was subsequently reduced by rotating the end-effector anti-clockwise in the opposite direction until the right fit around the coral was achieved. Fig. \ref{fig:demonstration}(c)(II) shows the coral's responsiveness to the continuously changing tension of the wire as the direction of rotation changes, which demonstrates its function as a physical twin to provide feedback of wire's tension, to prevent unintentional damage from overtightening.

The second demonstration explores the application of coral farming, whereby corals will need to be gently taken out of the tank for inspection or cleaning. To validate the coral's performance in water, a similar experimental setup was conducted as seen in Section \ref{section:experiments}, but with a 3-axis gantry system and water tank in place of the robotic manipulator, as in Fig. \ref{fig:demonstration}(d-e). The same gripper with the potted load cell was used, and the coral was placed in the base of the tank. The tank was subsequently filled with \SI{12}{\liter} of deionised water, submerging the coral completely. The same human guided force reference was used to determine the target grasping resistance. With the coral's resistance feedback, the gripper automatically stopped in time just before the target threshold is reached, as shown by Fig. \ref{fig:demonstration}(f)(II). Similarly to previous experiments, the fingertip forces showed large variation primarily caused by displacement of the coral due to the induced water movement by the gripper. In underwater environments, coral branches tend to passively sway in response to water currents. Robotic systems will need to account for these highly dynamic contact conditions, in addition to poor visibility inherent in underwater environments. Sensorising the object therefore provides an additional modality of data, which is critical to address the uncertainties in such tasks. Our method is promising to develop coral physical twins for underwater testing, as opposed to traditional methods of using pneumatic or hydraulic pressure-based sensors.  In our application, corals may need to be periodically removed from the water for inspection by farm handlers. Hence, these frequent transitions between air and water, along with buoyancy changes due to water depth, can affect sensor readings and may necessitate frequent recalibration. In contrast, the measurements provided by the LM corals are highly responsive, with minimal noise and drift, making them well-suited for manipulation tasks that require robust feedback, as demonstrated by the experiments. 

\section{Conclusion and Discussion}
\label{section:conclusion}

In this paper, we present a novel design and application of a soft sensing coral, developed through an automated design workflow that transforms the scans of real objects into functional sensing structures. Pathways for the LM channel were autonomously generated using waypoints defined by the user, allowing easy fabrication and sensorisation of geometrically challenging 3D morphologies, such as those seen in corals. The performance of the sensing coral was evaluated through a series of grasping experiments, which demonstrates sensitivity across all its branches and its reliability to be used as a tool to provide feedback for soft manipulation tasks. The sensing coral was developed as an additional sensing modality, to improve object representation during manipulation, with our results demonstrating its ability to capture intricate gripper-object interactions that cannot otherwise be detected with fingertip sensors alone. This is useful for delicate tasks that require gentle handling, such as labelling coral specimens or picking up corals for inspection and cleaning. Our work represents a crucial step toward enabling robots to support the cultivation of these endangered species, threatened by climate change. Beyond the demonstrated use case, our 3D printed soft sensors provide a generalised method to create sensorised and smart objects to train manipulation policies. The data collected will be made publicly available on CSIRO's Data Access Portal before the paper's final submission. Future work will focus on automating the integration of more complex sensor networks into arbitrary 3D geometrical structures and optimising the placement of these channels to allow for multi-contact localisation.  


%%%%%%%%%%%%%%%%%%%%%%%%%%%%%%%%%%%%%%%%%%%%%%%%%%%%%%%%%%%%%%%%%%%%%%%%%%%%%%%%

\section*{ACKNOWLEDGMENT}

The authors express their sincere gratitude to Russell Brinkworth for his valuable technical expertise, Tom Bridge and the Queensland Museum for the interview on coral sampling methods, and Stephen Rodan for allowing us to use the CHARM farming robot for submerged experiments.

%%%%%%%%%%%%%%%%%%%%%%%%%%%%%%%%%%%%%%%%%%%%%%%%%%%%%%%%%%%%%%%%%%%%%%%%%%%%%%%%

\bibliographystyle{IEEEtran}
\bibliography{references}

\end{document}